\documentclass[10pt,compsoc,peerreview]{IEEEtran}
\usepackage{times}
\usepackage{graphicx}
\usepackage{amsmath}
\usepackage{bm}
\usepackage{subeqnarray}
\usepackage{cases}
\usepackage{amssymb}
\newcounter{defcounter}
\usepackage{pdfpages}
\usepackage{color}

\usepackage{array,multirow}
\newenvironment{myequation}{%
	\addtocounter{equation}{-1}
	\refstepcounter{defcounter}
	
	\begin{equation}}
{\end{equation}}

\usepackage{latexsym}
\usepackage[utf8x]{inputenc}
\usepackage{amssymb,amsmath,amsfonts}
\usepackage{algorithm}
\usepackage{algorithmic}
\usepackage{lineno}
\usepackage{color}
\usepackage{url}
\usepackage{amsthm}

\newcommand{\cut}[1]{}

\begin{document}
	%
	\title{Output Constraint Transfer for Kernelized Correlation Filter in Tracking}
	%
	%
	%
	\author{Baochang Zhang$^1$, Zhigang Li$^1$, Xianbin Cao$^1$, Qixiang Ye$^2$, Chen Chen$^4$,  Linlin Shen*$^6$\footnote{correspondence ; llshen@szu.edu.cn}, Alessandro Perina$^3$, Rongrong Ji$^5$\\
		
		$^1$ School of Automation Science and Electrical Engineering,
		Beihang University, Beijing, China\\
		\and
		$^2$ University of Chinese Academy of Sciences, Beijing, China \\
		\and	
		$^3$ Microsoft Corporation, Redmond, WA, USA \\
		\and
		$^4$Center for Research in Computer Vision (CRCV) University of Central Florida, Orlando, FL, USA \\
		\and
		$^5$Xiamen University, Xiamen, China \\
		\and
		$^6$Shenzhen University, Shenzhen, China \\
	}
	\markboth{B.Zhang et al., Output Constraint Transfer for Kernelized Correlation Filter in Tracking}%
	{}
	
	\IEEEcompsoctitleabstractindextext{
		\begin{abstract}
			Kernelized Correlation Filter (KCF) is one of the state-of-the-art object trackers. However, it does not reasonably model the distribution of correlation response during tracking process, which might cause the drifting problem, especially when targets undergo significant appearance changes due to occlusion, camera shaking, and/or deformation. In this paper, we propose an Output Constraint Transfer (OCT) method that by modeling the distribution of correlation response in a Bayesian optimization framework is able to mitigate the drifting problem. OCT builds upon the reasonable assumption that the correlation response to the target image follows a Gaussian distribution, which we exploit to select {training samples} and reduce model uncertainty. OCT is rooted in a new theory which transfers data distribution to a constraint of the optimized variable, leading to an efficient framework to calculate correlation filters. Extensive experiments on a commonly used tracking benchmark show that the proposed method significantly improves KCF, and achieves better performance than other state-of-the-art trackers. To encourage further developments, the source code is made available { ~ https://github.com/bczhangbczhang/OCT-KCF ~ ;
				
		    [2]	Baochang Zhang, Z. Li, X. Cao, Qixiang Ye, C. Chen, L. Shen, A. Perina, and R. Ji,"Output Constraint Transfer for Kernelized Correlation Filter in Tracking," IEEE Transactions on Systems, Man, and Cybernetics:Systems, 2016, Digital Object Identifier 10.1109/TSMC.2016.2629509. } .
		\end{abstract}

	}

	%
	\maketitle
	
	\IEEEdisplaynotcompsoctitleabstractindextext

	%
	\IEEEpeerreviewmaketitle

	\section{Introduction}

	Visual object tracking is a fundamental problem in computer vision, which contributes to various applications including robotics, video surveillance, and intelligent vehicles \cite{1,2,ijcv}. While many works consider object tracking in simple scenes as a solved problem, on-line object tracking in uncontrolled real-world scenarios remains open, with key challenges like illumination change, occlusion, motion blur, and texture variation \cite{1,2,3,4}.
	To this end, the conventional data association and
	temporal filters \cite{Nummiaro03} that rely on motion modeling typically fail due to the dynamic and changing object/background appearances.

	Most recently, kernalized correlation filters (KCF), which aims to construct discriminative appearance model for  tracking from a learning-based perspective, has shown to be promising to handle the appearance variations \cite{kcf,stc,DSST}. KCF incorporates translated and scaled patches to make a kernelized model distinguishing between the target and surrounding environment \cite{kcf}. It also adopts Fast Fourier Transform (FFT) and Inverse FFT (IFFT) to improve the computational efficiency. Experimental comparisons show that KCF based tracking  is competitive among the state-of-the-art trackers in terms of speed and accuracy \cite{kcf}.  Although much success has been demonstrated, irregular correlation responses and target drifting have been observed. These are particularly common when updating target appearance in a long tracking stream with occlusion, camera shake and great appearance changes \cite{ma}. From the perspective of learning, sample noises are introduced to the filter, which degrades the
	model learning and drift the tracker away \cite{ma,wacv}. To alleviate such risk of drifting, we advocate that the tracker should model the correlation response (output) to reduce noisy samples to achieve stable tracking. We propose preventing the drifting through controlling maximum response to follow the Gaussian distribution, which not only reduce noise samples, but also gain the robustness to variations.
	
	As another intuition, it is well known that data lies on specific distributions, \emph{i.e.}, faces are considered to be from subspace \cite{J.Wright:PAMI2009,cvpr}. As long as the optimal solution resides on the data domain, the constraints derived from the data structure can bring robustness to the variations \cite{iconip,tld}.  To this end, any tracking framework taking advantage of the implicit data structure can improve tracking. Imposing a data structure (distribution) as a constraint is actually a new and flexible way to solve the optimization problems \cite{iconip,tld}, which has been promising in various learning algorithms. The main crux is how to efficiently embed structure constraint in the optimization method. In this paper, we demonstrate the existence of a highly practical solution to include Gaussian constraints in KCF.
	
\begin{figure*} [t!]
		\begin{center}
			\includegraphics[width=0.6\textwidth]{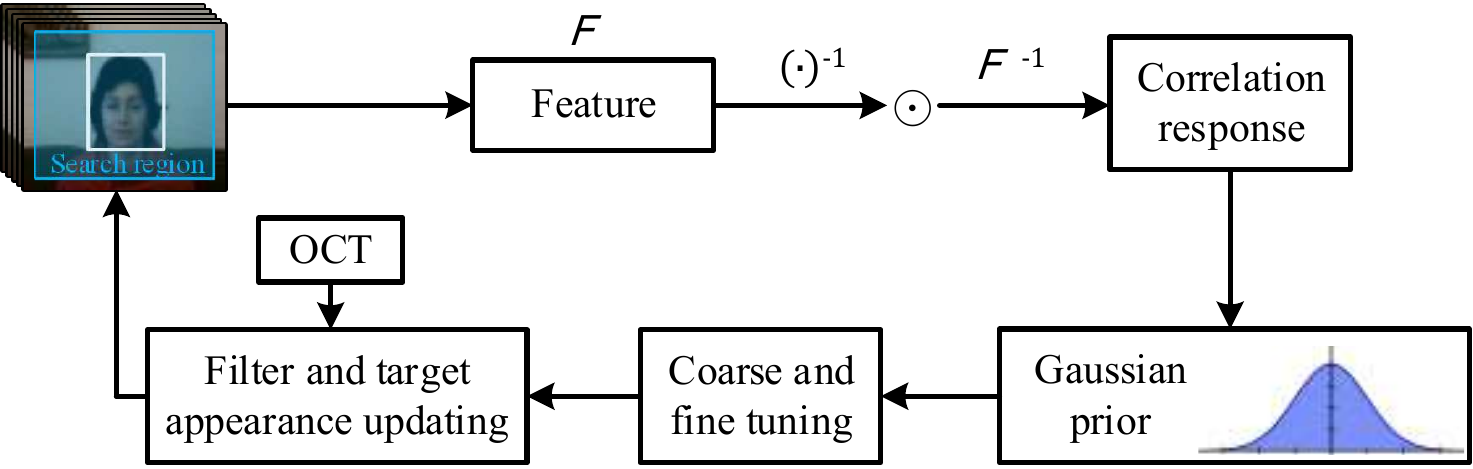}
		\end{center}
		\vspace{-0.1cm}
		\caption{A scheme of OCT-KCF for object tracking.}
		\vspace{-0.2cm}
		\label{fig:scheme}
	\end{figure*}

	{	Fig. \ref{fig:scheme} shows the proposed output constraint transfer (OCT) method\footnote{The source code will be publicly available on mpl.buaa.edu.cn.}, which mainly innovates at learning robust kernerlized correlation filters for object tracking. Two key innovations are introduced,
		(1) The Gaussian prior constraint is exploited to model the filter response and reduce noisy samples,
		(2) A new theory termed OCT is proposed to transfer data distribution to be a constraint of the optimized variable.
		By the constraint, our correlation filters are particularly prone to find the \textit{data} (response output) following a certain distribution \footnote{Assumed Gaussian for its simplicity, although other complex distributions may be more reasonable} and gain the robustness to variations. By the proposed OCT theory, instead of directly controlling the response output in a brute-force way, we alternatively transfer the distribution information from the data to be a constraint of the optimized variable.}
	
	The Gaussian  assumption on correlation output is supported from three aspects. (1) It is first supported by \cite{kcf}, and shows that a single threshold on the correlation response (output) is used, which inspire us that the correlation response (output) actually follows a simple distribution, \emph{i.e.}, Gaussian.
	(2) As evident on tracking, a simple distribution is necessary and significant to achieve high efficiency.  The complex distribution, \emph{i.e.}, Gaussian mixture model, can not result in an efficient model as ours.  As for the complex distribution, it would be further considered in our future work and could be possible to have a more general theory.
	(3) As a final evidence, our extensive experiments on the commonly used benchmark \cite{dataset} confirm that the Gaussian distribution is highly effective.

	
	The rest of the paper is organized as  follows: Section 2 introduce the related work. We present the constraint problem on correlation filters in Section 3, and detail how Gaussian constraints can be efficiently embedded in an online optimization framework in Section 4. Finally, extensive experiments are discussed in Section 5, while we draw our conclusions in Section 6 .
	
	\section{Related work}
	Visual tracking has been extensively studied in the literature \cite{dataset,wacv}. In this section, we discuss the methods closely related to this work, \emph{i.e.}, the appearance models, and more particularly, the correlation filter based models.
	
	An appearance model consists of learning a classifier online, to predict the presence or absence of the target in an image patch. This classifier is then tested on
	many candidate patches to find the most likely location \cite{kcf,z.han,Kwon,whu}. Popular learning schemes include kernel learning \cite{svm,trakcing_adaptive}, latent structure \cite{latent}, multiple instance learning \cite{wmil}, boosting \cite{boost1,boost2}, metric learning \cite{metric} and structured learning \cite{struck}. However, the online tracking algorithms often encounter the drifting problems. As for the self-taught learning, these misaligned samples are likely to be added and degrade the appearance models. To avoid drifting, the most famous Tracking-By-Detection (TLD) method employs positive-negative (P-N) learning to choose ``safe" samples \cite{tld}.
	The Compressed Tracking method employs non-adaptive random projections that preserve the structure of the image feature space of objects \cite{ct}. It compresses samples of foreground targets and the background using the same sparse measurement matrix to guarantee the stability of tracking \cite{ct}.
	
	The initial motivation for our research was the recent success of correlation filters in tracking \cite{MOSSE}.
	Correlation filters have been proved to be competitive with far more complicated approaches, but using only a fraction
	of the computational power, at hundreds of frames per second. They take advantage of the fact that the convolution
	of two patches is equivalent to an element-wise product in the FFT domain. Thus, by formulating their
	objective in the FFT domain, they can specify the desired output of a linear classifier for several translations,
	or image shifts \cite{kcf}.	
	Taking the advantages of correlation filters, Bolme \textit{et al.} propose to learn a minimum output sum of squared error (MOSSE) \cite{MOSSE} filter for visual tracking on gray-scale
	images.
	Heriques \textit{et al.} propose using correlation filters in a kernel
	space based on CSK \cite{csk}, which achieves the highest
	speed in the commonly used benchmark \cite{dataset}.  CSK is introduced based on kernel ridge regression, which has been one of the hottest topics in correlation filter learning. Using a dense sampling strategy, the circulant structure exploits data redundancy
	to simplify the training and testing process. By using HOG features, KCF is further proposed to improve the performance of CSK.
	In \cite{Danelljan}, Danelljan \textit{et al.} exploit the color attributes of a target
	object and learn an adaptive correlation filter by mapping
	multi-channel features into a Gaussian kernel space. Recently, Ma \textit{et al.} introduce a re-detecting process to further improve the performance of KCF \cite{kcf}.
	Zhang \textit{et al.} \cite{stc} incorporate context information
	into filter learning and model the scale change based on consecutive
	correlation responses. The DSST tracker \cite{DSST} learns
	adaptive multi-scale correlation filters using HOG features
	to handle the scale variations. Recent works involve using learned Convolutional Filters for visual object tracking [33, 35, 36].
	Although much success has been demonstrated,  the existing works do not principally incorporate the distribution information into the procedure of solving the optimized variable.
	
	\section{Output Constraint Transfer in KCF}
	In this section we first introduce KCF, and then describe how the response output is constrained by a Gaussian distribution.
	\subsection{Kernelized correlation filter}
	KCF starts from the kernel ridge regression method \cite{kcf}, which is formulated as:
	\begin{myequation}
		\label{p1}
		\begin{aligned}
			&\underset{w,\xi}{\min}
			& &\sum_{i}\xi_i^2 \\
			&\textit{subject to}
			& & y_i-\bm{w^T\phi}(x_i)=\xi_i     ~~\forall{i};
			& & ||w||\le B,&
		\end{aligned}
	\end{myequation}
	where $x_i$ is the $M\times N$-sized image. $\phi(.)$ is a non-linear transformation. $\phi(x_i)$ (later $\phi_i$) and $y_i$ are the input and output, respectively.  $\xi_i$ is a slack variable.  $B$ is a small constant. 
	Based on the Lagrangian method, the objective  corresponding to \ref{p1} is rewritten as:
	
	\begin{equation}
	\begin{aligned}
	&\mathcal{L}_p=\sum_{i=1}^{M\times N}\xi_i^2+\sum_{i=1}^{M\times N}\beta_i \left[ y_i-\bm{w}^T\phi_i-\xi_i \right]+\lambda  (\left\| \bm{w} \right\|^2-B^2),
	\label{equp1}
	\end{aligned}
	\end{equation}
	where $\lambda$ is  a regularization parameter ($\lambda\ge0$). 
	From Equ. \ref{equp1}, we have:
	\begin{equation}
	\begin{aligned}
	&\bm{\alpha}=\left(K+\lambda \bm{I}\right)^{-1}\bm{y},\\
	&\bm{w}=\sum_{i}\alpha_{i}\phi_i.
	\label{e1}
	\end{aligned}
	\end{equation}    %
	The matrix $K$ with elements $K_{ij}=k(P^i\bm{x},P^j \bm{x})$  is  circulant given a kernel such as the gaussian kernel $k$  \cite{csk}.
	Taking advantage of the circulant matrice, the FFT of $\bm{\alpha}$ denoted by $\mathcal{F}(\bm{\alpha})$ is calculated by:
	\begin{equation}
	\begin{aligned}
	\mathcal{F}(\bm{\alpha})=\frac{\mathcal{F}(y)}{\mathcal{F}(k^{xx})+\lambda},
	\label{e2}
	\end{aligned}
	\end{equation}
	where $\mathcal{F}$ denotes the discrete Fourier operator, and $k^{xx}$ is the first row of the circulant  matrix $K$. In tracking, all candidate patches that are cyclic shifts of test patch $z$ are evaluated by:
	\begin{equation}
	\begin{aligned}
	\mathcal{F}(\hat{\bm{y}})=\mathcal{F}(k^{z\hat{x}})\odot\mathcal{F}(\bm{\alpha}),
	\label{e4}
	\end{aligned}
	\end{equation}
	where $\odot$ is the element-wise product and $\hat{x}$ is a learned target appearance image calculated by Equ. 6a \cite{ma}, $\mathcal{F}(\hat{\bm{y}})$ is the output response for all the testing patches in frequency domain. We then have:
	\begin{equation}
	\begin{aligned}
	\hat{\bm{y}}= \max (\mathcal{F}^{-1}({{\mathcal{F}(\hat{\bm{y}})}})),
	\end{aligned}
	\label{e5}
	\end{equation}
	where $\mathcal{F}^{-1}$ is the inverse FFT. The target position is the one with the maximal value among $\hat{\bm{y}}$ calculated by Eq. 5. The target appearance and
	correlation filter are then updated with a learning rate $\eta$ as:
	\begin{subequations}
		\begin{numcases}{}
		\hat{x}^t=(1-\eta)\hat{x}^{t-1}+\eta x^t,
		\label{e6a}\\
		\mathcal{F}(\bm{\alpha^t})=(1-\eta)\mathcal{F}(\bm{\alpha^{t-1}})+\eta\mathcal{F}(\bm{\alpha}).\label{e6b}
		\end{numcases}
	\end{subequations}
	Kernel ridge regression relies on computing kernel
	correlation ($k^{xx}$ and $k^{z\hat{x}}$). Considering that
	kernel correlation consists of computing the kernel for all
	relative shifts of two input vectors. This represents the
	last computational bottleneck, as an evaluation
	of $n$ kernels for signals of size $n$ will have quadratic
	complexity. However, using the cyclic shift model will
	allow us to efficiently exploit the redundancies in this
	expensive computation. The computational complexity for the full kernel correlation
	is only $O(n \cdot log(n))$.
	\subsection{Problem formulation}
	In tracking applications, the  correlation response of the target object is assumed to follow a Gaussian distribution, which is not discussed in the existing works. In this section, we solve the \ref{p1} by exploiting the Gaussian assumption in an optimization process. Now, the original problem \ref{p1} can be rewritten in the $t^{th}$ frame as:
	\begin{myequation}
		\begin{aligned}
			&\underset{w,\xi}{\min}
			& \sum_{i}\xi_i^2 \\
			&\textit{s.t.:}\\
			& & y_i-{\bm{w}^{T,t}}\bm{\phi}_i=\xi_i,\\
			& & \bm{\hat{y}}^t \thicksim \mathcal{N}(\mu^t,\sigma^{2,t}), \\
			& & ||w||\le B,&\
			\label{p2}
		\end{aligned}
	\end{myequation}
	where $y_i$ is the Gaussian function label for the $i^{th}$ sample $\bm{\phi}_i$ in the $t^{th}$ frame \cite{kcf}. $\mu,\sigma^{2}$ are the mean and variance of the Gaussian model $\mathcal{N}$ respectively. $\hat{y}^t$ is a new variable to represent the response of the target image based on $\bm{w}^{T,t}$ and Eq.5. As mentioned above, Gaussian prior is defined as:
	\begin{equation}
	\bm{\hat{y}}^t \thicksim \mathcal{N}(\mu^t,\sigma^{2,t}).
	\end{equation}
	In Problem \ref{p2}, only
	$ \bm{\hat{y}}^t \thicksim \mathcal{N}(\mu^t,\sigma^{2,t})$ is unsolved. As shown in the maximum likelihood method in the probability theory \cite{ml}, Gaussian prior can be alternatively solved through minimizing
	$\frac{({\bm{\hat{y}}^t} - {\mu^t})^2}{2\sigma^{t}} + \frac{ln(2\pi \cdot \sigma^{t})}{2}$. As only $\frac{({\bm{\hat{y}}^t} - {\mu^t})^2}{2\sigma^{t}}$   is related to the optimized variable, $\mu^t$ and $\sigma^{2,t}$ are solved iteratively. And for simplicity, $\sigma^{2,t}$ can be considered as a constant in the ${t}^{th}$ time. Thus, the denominator is ignored, we alternatively minimize $({\bm{\hat{y}}^t}-u^t)^2$. The smaller the value is, the more possible the correlation response in current frame satisfies the Gaussian prior.  \footnote{$\mu$ and $\sigma^2$  are calculated based on all previous frames in the tracking procedure.}
	
	\section{ OCT based KCF}
	In the previous section, we describe a new framework to calculate kernelized correlation filter. Nevertheless, it remains complex to solve the tracking problem, due to the new variable $\bm{\hat{y}}^t$. Here we introduce the proposed OCT theory to further reformulate  \ref{p2} into an extremely simple problem.
	
	\subsection{Theory of transferring constraints: OCT}
	The OCT theory aims to simplify the optimization process in particular for  $({\bm{\hat{y}}^t}-u^t)^2$. As a result of the theory  $\bm{\hat{y}}^t$ is replaced by a new constraint only added on the variable $\bm{w}$. This is remarkable, the Gaussian constraint is deployed without extra complexity, \emph{i.e.}, $\bm{\hat{y}}^t$ is not involved. Here $\bm{\hat{y}}^t = \bm{w}^{T,t}x^t$ with $x^t$ as the target object.\\ 
	$\textbf{Theorem}$ \textit{Minimizing of $(\bm{w}^{T,t}x^t - \mu^t)^2$ is transfered to  minimizing  $||\bm{w}^t-\bm{w}^{t-1}||^2$,  when the learned target appearances have no great changes in two consecutive frames.}
	
	{Based on the theorem, the data distribution is transfered to a constraint only for the unsolved variable, by which Problem \ref{p2} is further relaxed, leading to an  extremely efficient method to calculate correlation filters.}\\
	$\textbf{Proof}$.
	The mean of Gaussian is updated as:
	\begin{equation}
	\mu^{t}=(1-\rho)\mu^{t-1}+\rho {\bm{w}^{T,t}}{\hat{x}^t},
	\label{e8}
	\end{equation}
	where $\hat{x}^t$ is the learned target appearance  in the $t^{th}$ frame, iteratively acquired by:
	\begin{eqnarray}
	\hat{x}^t=(1-\rho)\hat{x}^{t-1}+\rho x^t,
	\label{e9}
	\end{eqnarray}
	where $x^t$ is the target in the $t^{th}$ frame.
	Similar to Eq.\ref{e8},  $\mu^{t-1}$ can be calculated as:
	\begin{equation}
	\mu^{t-1}=(1-\rho)\mu^{t-2}+\rho {\bm{w}^{T,t-1}}{\hat{x}^{t-1}}.
	\label{e10}
	\end{equation}
	By plugging  Eq.\ref{e10}  back into Eq.\ref{e8}, we get:
	\begin{eqnarray}
	\mu^{t}=(1-\rho)((1-\rho)\mu^{t-2}+\rho {\bm{w}^{T,t-1}}{\hat{x}^{t-1}})+
	\rho {\bm{w}^{T,t}}\hat{x}^t,
	\label{e11}
	\end{eqnarray}
	which is rewritten as:
	\begin{align}
	{\bm{w}^{T,t}}{x^{t}}-\mu^{t}=&-\rho(1-\rho) {\bm{w}^{T,t-1}}{\hat{x}^{t-1}}-
	(1-\rho)^2\mu^{t-2}-\nonumber \\&\rho  {\bm{w}^{T,t}}{\hat{x}^{t}}+{\bm{w}^{T,t}}{x^{t}}.
	\label{e12}
	\end{align}
	Here $\bm{w}^{T,t}x^t$ is approximated by $\bm{w}^{T,t}\hat{x}^t$, which does not change the tracking result. Thus, Eq.\ref{e12} is rewritten as :
	\begin{align}
	{ \bm{w}^{T,t}}{x^{t}}-\mu^{t}=&-\rho(1-\rho) {\bm{w}^{T,t-1}}{\hat{x}^{t-1}}-
	(1-\rho)^2\mu^{t-2}+ \nonumber \\& (1- \rho){\bm{w}^{T,t}}{\hat{x}^t}+\epsilon_1.
	\end{align}
	where $\epsilon_1$ is a small constant. Plugging Eq.\ref{e9} back into the above equation, we have:
	\begin{align}
	{ \bm{w}^{T,t}}{x^{t}}-\mu^{t}=&-\rho(1-\rho) {\bm{w}^{T,t-1}}{\hat{x}^{t-1}}-
	(1-\rho)^2\mu^{t-2} \;\nonumber \\
	&+(1-\rho)^2{\bm{w}^{T,t}}{\hat{x}^{t-1}}+
	(1-\rho)\rho{\bm{w}^{T,t}}{\hat{x}^{t}}	
	+\epsilon_1.
	\end{align}
	Based on the hypothesis that the learned target appearances ($\hat{x}^t$, $\hat{x}^{t-1}$) have no great changes in two consecutive frames, we have:
	\begin{align}
	{\bm{w}^{T,t}}{x^{t}}-\mu^{t}=&\rho(1-\rho)(\bm{w}^{T,t}-{\bm{w}^{T,t-1}})\hat{x}^{t-1}\;\nonumber\\
	&+(1-\rho)^2({\bm{w}^{T,t}}\hat{x}^{t-1}-\mu^{t-2})+\epsilon_2,
	\label{e14}
	\end{align}
	where  $\epsilon_2$ is a small constant.
	\begin{align}
	||{\bm{w}^{T,t}}{x^{t}}-\mu^{t}||&\le||(1-\rho)^2({\bm{w}^{T,t}}\hat{x}^{t-1}-\mu^{t-2})||\,\nonumber\\
	&+||\rho(1-\rho)({\bm{w}^{T,t}}-{\bm{w}^{T,t-1}})\hat{x}^{t-1}||
	+||\epsilon_2||\;\nonumber\\
	&\le\rho(1-\rho)||\bm{w}^t-\bm{w}^{t-1}||\cdot||\hat{x}^{t-1}||+C,
	\end{align}
	where $C$ is a constant. From the above inequality, the minimization of  ${(\bm{w}^{T,t}}x^t-\mu^t)^2$ is converted to minimizing $||\bm{w}^t-\bm{w}^{t-1}||^2$, \textbf{Theorem} is proved.$\Box$
	\subsection{The OCT solution to the problem  \ref{p2}}
	Bayesian optimization is a powerful framework which has been successfully applied to solve various problems, \emph{i.e.}, parameter tuning. The Bayesian optimization can also be used to solve Problem \ref{p1}.
	%
	Two of the KKT conditions from Equ.1 are:
	\begin{equation}
	2\xi_i=\beta_i^t.\text{~~}
	\end{equation}
	\begin{equation}
	2\lambda \bm{w}=\sum_{i}\beta_i^t\bm{\phi}_i.
	\label{eqw}
	\end{equation}
	According to our theory, we add the minimizing of $||\bm{w}^t-\bm{w}^{t-1}||^2$ to replace the Gaussian constraint in Problem \ref{p2}. However, it is still a little complicated for our problem. Based on Eq.\ref{eqw}, we simply use $||\bm{\beta}^t-\bm{\beta}^{t-1}||^2 $, and obtain a dual form for Problem \ref{p2} via the Lagrangian method, which is formulated in a Bayesian framework as:
	\begin{align}
	\mathcal{L}_P(\alpha|(\mu^t,\sigma^{2,t}))=&-\frac{1}{4}\sum_{i}{\beta_i^{2,t}}-\frac{1}{4\lambda}\sum_{i,j}\beta_i^t\beta_j^t K_{ij}\;\nonumber
	\\&+\sum_{i}\beta_i^t y_i-s\sum_{i}(\beta_i^t-\beta_i^{t-1})^2-\lambda B^2.
	\end{align}
	%
	Redefining $
	\alpha_i^{t}=\beta_i^t/2\lambda\text{~}$, we come up with the following optimization problem:
		\begin{align}
	\underset{\bm{\alpha},\lambda}{\max}       ~~-&\lambda^2\sum_i{\alpha_i^{2,t}}+2\lambda\sum_{i}\alpha_i^ty_i
	-\lambda\sum_{i,j}\alpha_i^t\alpha_j^tK_{ij}- \nonumber\\& 4\lambda^2s\sum_{i}(\alpha_i^t-\alpha_i^{t-1})^2.
	\end{align}
	and we have:
	\begin{equation}
	(\lambda I+4\lambda sI+\mathrm{K})\bm{\alpha}^t=\bm{y}+4\lambda s\bm{\alpha}^{t-1}.
	\end{equation}
	Then the FFT of $\bm{\alpha}$ is calculated as:
	\begin{equation}
	F(\bm{\alpha}^t|(\mu^t,\sigma^{2,t}))=\frac{F(\bm{y})+4\lambda s F(\bm{\alpha}^{t-1})}{F(\bm{k})+\lambda+4\lambda s},
	\end{equation}
	which is rewritten as:
		\begin{align}
	F(\bm{\alpha}^t|(\mu^t,\sigma^{2,t}))=&\frac{F(\bm{k})+\lambda}{F(\bm{k})+\lambda+4\lambda s}\odot\frac{F(\bm{y})}{F(\bm{k})+\lambda}
	+\nonumber\\& \frac{4\lambda s}{F(\bm{k})+\lambda+4\lambda s}\odot F(\bm{\alpha}^{t-1}).\nonumber\\
		\end{align}
	Defining $\bm{\eta}$ as:
	\begin{equation}
	\bm{\eta}=\frac{F(\bm{k})+\lambda}{F(\bm{k})+\lambda+4\lambda s},
	\end{equation}
	we have:
	\begin{equation}
	F(\bm{\alpha}^t|(\mu^t,\sigma^{2,t}))=\bm{\eta}\odot F(\bm{\alpha})+(1-\bm{\eta})\odot F(\bm{\alpha}^{t-1}),
	\label{e24}
	\end{equation}
	where $\mu^t,\sigma^{2,t}$ are used to select the samples as shown in Eq.26. $\bm{\eta}$  is a matrix with the same size as $F(\bm{\alpha})$. According to Eq.\ref{e24}, the update of the filter relies on the evolving $\bm{\eta}$, which is different with KCF (Eq.\ref{e6b}) that relies on a constant. More details about $\bm{\eta}$ can also refer to our source code.  To be concluded from the results mentioned above, the iterative formula of correlation filter  Eq.\ref{e24} is obtained from the theoretical derivation.

	\subsection{Coarse and fine tuning based on Gaussian prior}
	Due to the appearance variations of the target, the tracker might gradually drift and finally fail. Different from existing works using  threshold to detect the failure case, we argue that the property of Gaussian prior can well prevent drifting. In particular, we adopt the Gaussian prior to select samples when their response output belong to a Gaussian distribution, that is:
	%
	The sample is chosen, only when its response output belongs to a Gaussian distribution:
	\begin{eqnarray}
	\label{drifting}
	\left|\frac{{\bm{\hat{y}}^t}-\mu^t}{\sigma^t}\right|\ < \mathcal{T}_g,
	\end{eqnarray}
	where $\mathcal{T}_g = 1.6$ is empirically set to a constant. Here we introduce a fine-tuning process to precisely localize the target for sample selection in a local region, instead of searching over the whole image extensively. The tracker activates the fine-tune process when the maximal correlation response is out of the Gaussian distribution (drifting).
	%
	%
	We first detect the coarse region where the target is most likely to appear near the location in previous frame. We then search a coarse region from $n_t$ directions around the center of the latest location $(x_0,y_0)$. 
	The coordinates of a center location for coarse regions are calculated by:
	\begin{equation}
	\begin{aligned}
	p_x=\begin{cases}
	x_0+i_r*r_s*cos(i_t*t_s) & \text{for $i_t$ mod 2=0}\\
	x_0+i_r*r_s*cos(i_t*t_s+\phi) & \text{for $i_t$ mod 2=1},
	\end{cases}
	\end{aligned}
	\label{e26}
	\end{equation}
	\begin{equation}
	\begin{aligned}
	p_y=\begin{cases}
	y_0+i_r*r_s*sin(i_t*t_s) & \text{for $i_t$ mod 2=0}\\
	y_0+i_r*r_s*sin(i_t*t_s+\phi) & \text{for $i_t$ mod 2 =1}.
	\end{cases}
	\end{aligned}
	\label{e27}
	\end{equation}
	\begin{algorithm} \caption{\textbf{- Output constraint transfer algorithm for object tracking}}
		\label{alg:proposed tracking algorithm}
		\begin{algorithmic}[1]
			\STATE  Initial target bounding box $\bm{b}_0=[x_0,y_0,w,h] $,
			\STATE \textbf{if} the frame $n\le 7$
			\REPEAT
			
			\STATE Crop out the search windows  according to $\bm{b}_{n-1}$, and extract the HOG features.
			\STATE Compute the maximum correlation response  $\hat{\bm{y}}$  using Eq.\ref{e4} and Eq.\ref{e5} and record the maximal correlation response as ${y_{n}}$
			\STATE The position is obtained according to the maximal correlation response
			\STATE Updating target appearance  and correlation filter  using Eq.\ref{e6a} and Eq.\ref{e24}.
			\UNTIL  $n==7$
			\STATE \textbf{end}
			\STATE Compute the mean $\mu$ and variance $\sigma^2$ using all previous frames.
			\STATE \textbf{if} $n>7$
			\REPEAT
			\STATE  Crop out the search window  and extract the HOG features.
			\STATE Compute the maximal correlation response  $\hat{\bm{y}}$ using Eq.\ref{e4} and Eq.\ref{e5}.
			\STATE \textbf{if} $
			\left|\frac{{\bm{\hat{y}}}-\mu}{\sigma}\right|\ > \mathcal{T}_g
			$
			\STATE Crop out the coarse regions\\  $Z=\{z_1,z_2,...,z_{n_r*n_t}\}$ according to the coordinates calculated by Eq.\ref{e26} and Eq.\ref{e27} around the center of $\bm{b}_{n-1}$
			\STATE \textbf{Coarse searching step}:\\Detect the patch $\hat{z}$ in which the target appears with maximal probability using Eq.\ref{e29} and Eq.\ref{e30}
			\STATE \textbf{Fine searching step}:\\
			Locate the object precisely  using Eq.(\ref{e4}) and
			Update target appearance and correlation filter using Eq.\ref{e6a} and Eq.\ref{e24}
			\STATE \textbf{end}
			\STATE Updating $\mu$ and $\sigma^2$
			\UNTIL End of the video sequence.
			\STATE \textbf{end}
		\end{algorithmic}
	\end{algorithm}
	where $r_s=\frac{radius}{n_r},i_r\in\{1,...,n_r\},t_s=\frac{2\pi}{n_t},i_t\in \{1,...,n_t\},\phi=\frac{t_s}{2} $
	. Finally, $n_r*n_t$ patches centered around the target are cropped as:
	\begin{equation}
	Z=\{z_1,z_2,,,z_{n_r*n_t}\}.
	\end{equation}
	In the coarse process, the maximal correlation response of each patch is obtained by:
	\begin{equation}
	\begin{aligned}
	r_i=max(F^{-1}(\mathcal{F}(z_i)=\mathcal{F}(k^{z_i\hat{x}})\odot\mathcal{F}(\bm{\alpha}))).
	\end{aligned}
	\label{e29}
	\end{equation}
	Then the patch in which the target appears with maximum probability is calculated as:
	\begin{equation}
	\begin{aligned}
	\hat{z}=\underset{{i}}arg \max(z_1,...,z_i,...z_{n_r*n_t}).
	\end{aligned}
	\label{e30}
	\end{equation}
	
	The fine-tuning step is executed to find the location ($\hat{z}$) of the object precisely as shown in Eq.(\ref{e4}) .
	The initialized process is empirically set during the first 20 frames. The fine-tuning strategy is easily implemented to update the localization of the tracked target. To sum up, Algorithm 1 recaps the complete method.

	\section{Experiments}
	In this section, we evaluate the performance of our tracker on 51 sequences of the commonly used  tracking benchmark \cite{dataset}. In this tracking benchmark \cite{dataset}, each sequence is manually tagged with 11 attributes which represent challenging aspects in visual tracking, including \emph{illumination variations}, \emph{scale variations}, \emph{occlusions}, \emph{deformations}, \emph{motion blur}, \emph{abrupt motion}, \emph{in-plane rotation}, \emph{out-of-plane rotation}, \emph{out-of-view}, \emph{background clutters} and \emph{low resolution}.
	
\begin{figure} [t!]
			\begin{center}
				\includegraphics[width=0.5\textwidth]{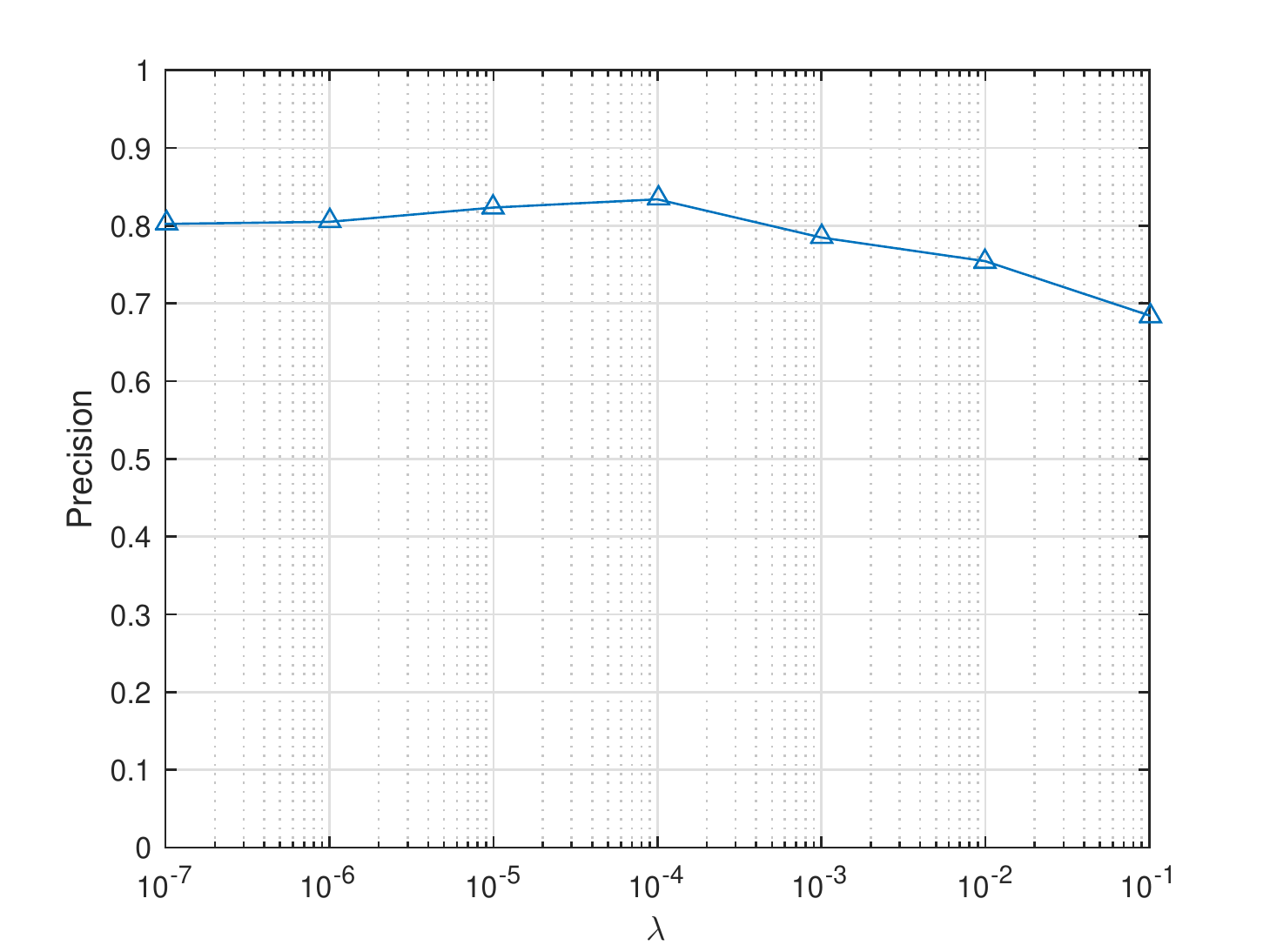}
			\end{center}
			\vspace{-0.1cm}
			\caption{The evaluation of $\lambda$ based on precision.}
			\vspace{-0.2cm}
			\label{fig:lambda}
		\end{figure}

\begin{figure} [t!]
			\begin{center}
				\includegraphics[width=0.5\textwidth]{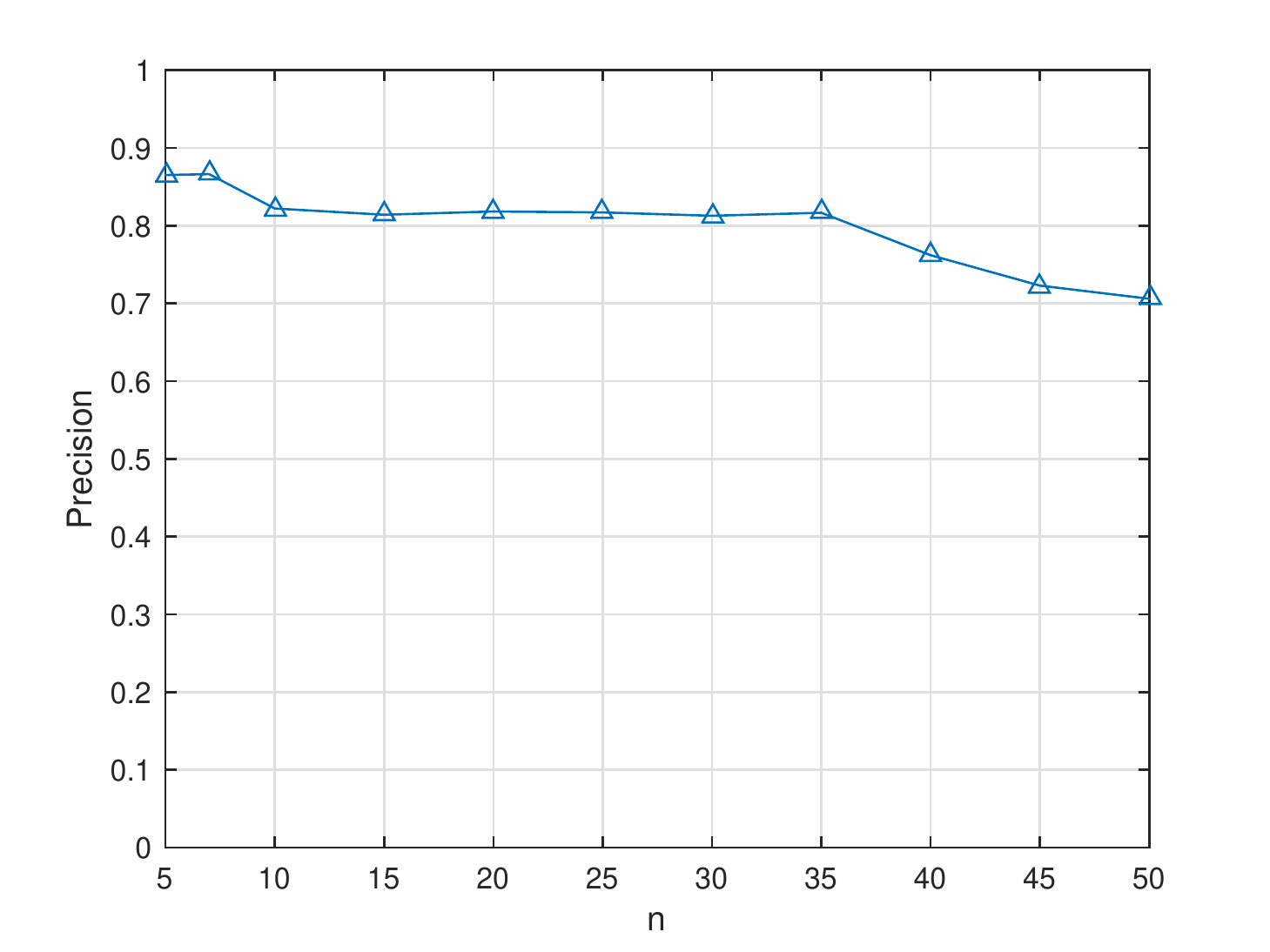}
			\end{center}
			\vspace{-0.1cm}
			\caption{The evaluation of $n$ based on precision.}
			\vspace{-0.2cm}
			\label{fig:frames}
	\end{figure}

\begin{figure} [t!]
			\begin{center}
				\includegraphics[width=0.5\textwidth]{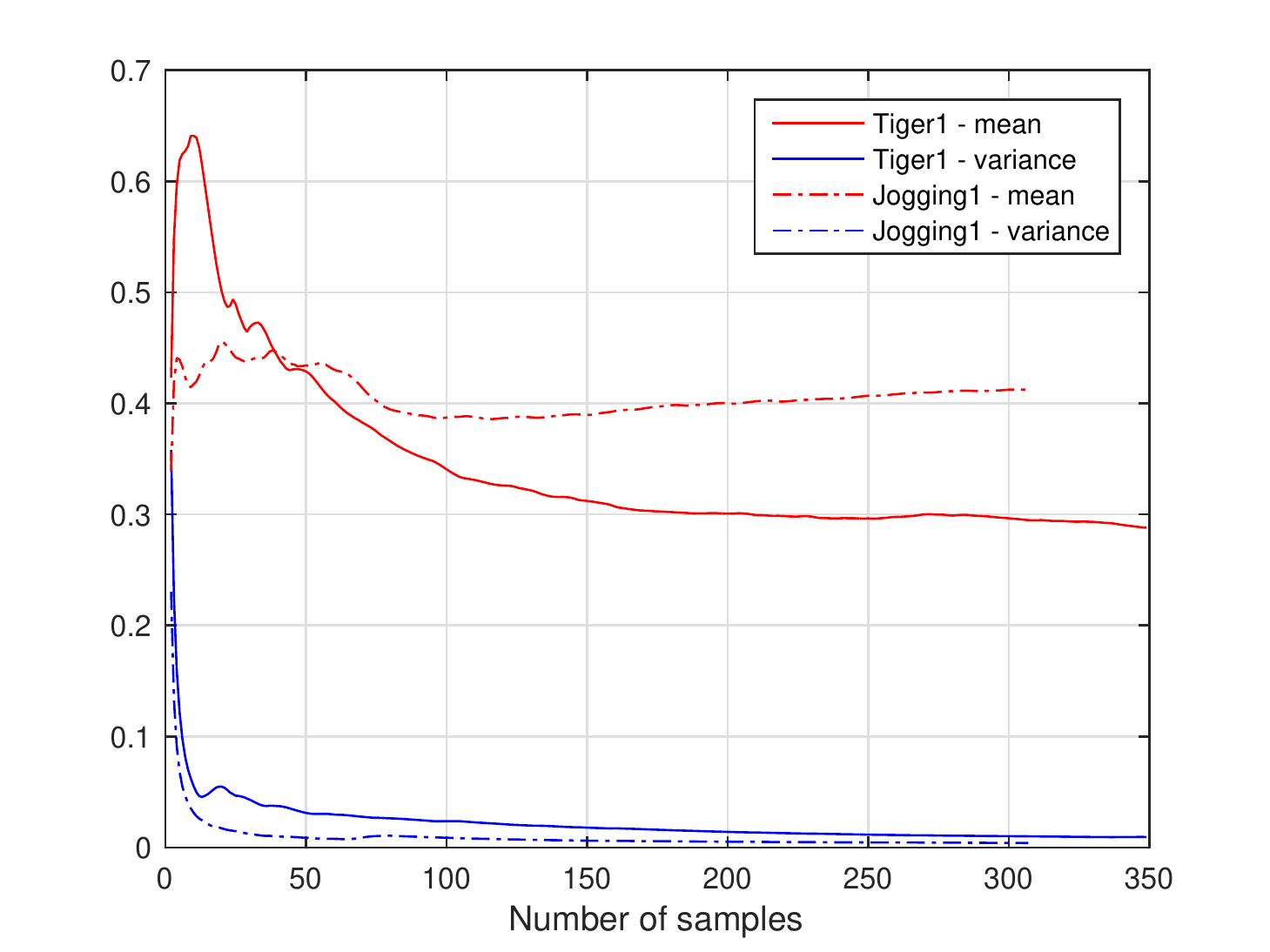}
			\end{center}
			\vspace{-0.1cm}
			\caption{The illustration of Gaussian mean and variance on the Tigger1 and Jogging1 sequences.}
			\vspace{-0.2cm}
			\label{fig:gaussian1}
		\end{figure}

\begin{figure} [t!]
		\begin{center}
			\includegraphics[width=0.5\textwidth]{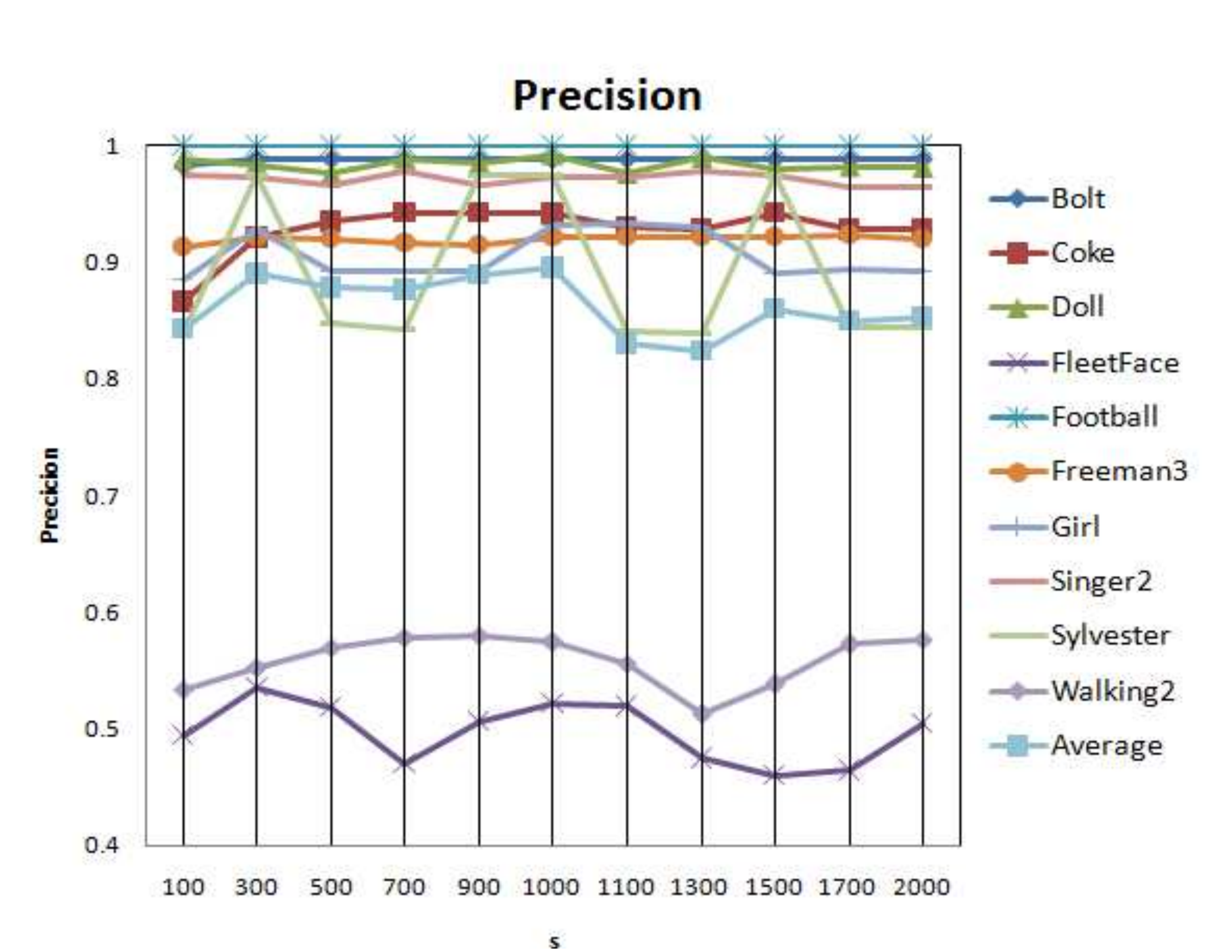}
		\end{center}
		\vspace{-0.1cm}
		\caption{The evaluation of $s$ based on precision.}
		\vspace{-0.2cm}
		\label{fig:ss}
	\end{figure}

	\textbf{Parameters evaluation:}  {We have tested the robustness of the proposed method in various parameter settings. For example, an experiment is done based on a subset of \cite{dataset}\footnote{The datasets are shown in Fig. \ref{fig:ss}}  as shown in Fig. \ref{fig:lambda}, the precision is not changed much when $\lambda$ is set from $10^{-7}$ to $10^{-3}$. About the initialized number of samples for Gaussian model, we tested different values in Fig. \ref{fig:frames}, and the performance is very stable around 20 that is finally chosen in the following experiments. Moreover, we illustrate Gaussian mean and variance in the tracking process  in Fig. \ref{fig:gaussian1}, which appear to be stable if the target is well tracked for the Tiger1 sequence, otherwise it seems randomly for the  Jogging1 sequence due to the wrong candidate  tracked. }  $s$ is a parameter used in Equ. \ref{e24}, the experiment based on a subset of \cite{dataset} is done as shown in Fig. \ref{fig:ss}. The performance of OCT is affected a litter by choosing different values of $s$. On most sequences (also average) the results on $s=1000$ is better than others. So we choose $s=1000$ in our experiment.  To be consistent with \cite{kcf}, we set $\lambda =10^{-4}$, $\rho=\frac{1}{t}$with $t$ as the frame number, and the searching size is 1.5. The Gaussian kernel function (standard variance = $0.5$) and most parameters used in OCT-KCF are empirically chosen according to \cite{kcf}. For other parameters, we empirically set  $n_r =5$, $n_t =16$ on all sequences.  \\  

	\begin{figure*} [t!]
		\begin{center}
			\includegraphics[width=1\textwidth]{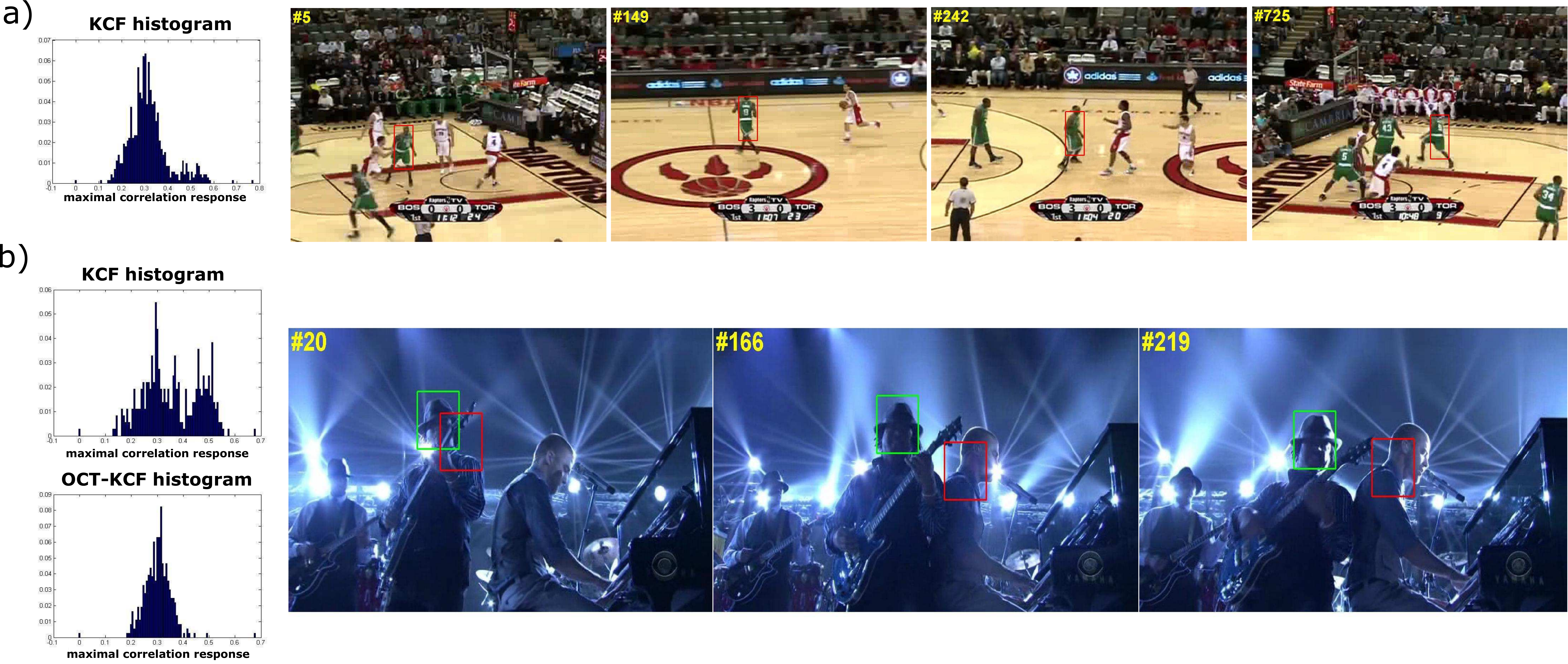}
		\end{center}
		\vspace{-0.1cm}
		\caption{Illustration of KCF and OCT-KCF on basketball and shaking sequences. a) A good performance is achieved when the response(output) of KCF is observed to follow a Gaussian distribution on the basketball sequence. b) OCT (green rectangular) is used to improve the performance of KCF (red rectangular) on the shaking sequence, and correlation response in OCT-KCF follows a Gaussian distribution.}
		\vspace{-0.2cm}
		\label{fig:gaussian}
	\end{figure*}
	
	\begin{figure*} [t!]
		\begin{center}
			\includegraphics[width=0.9\textwidth]{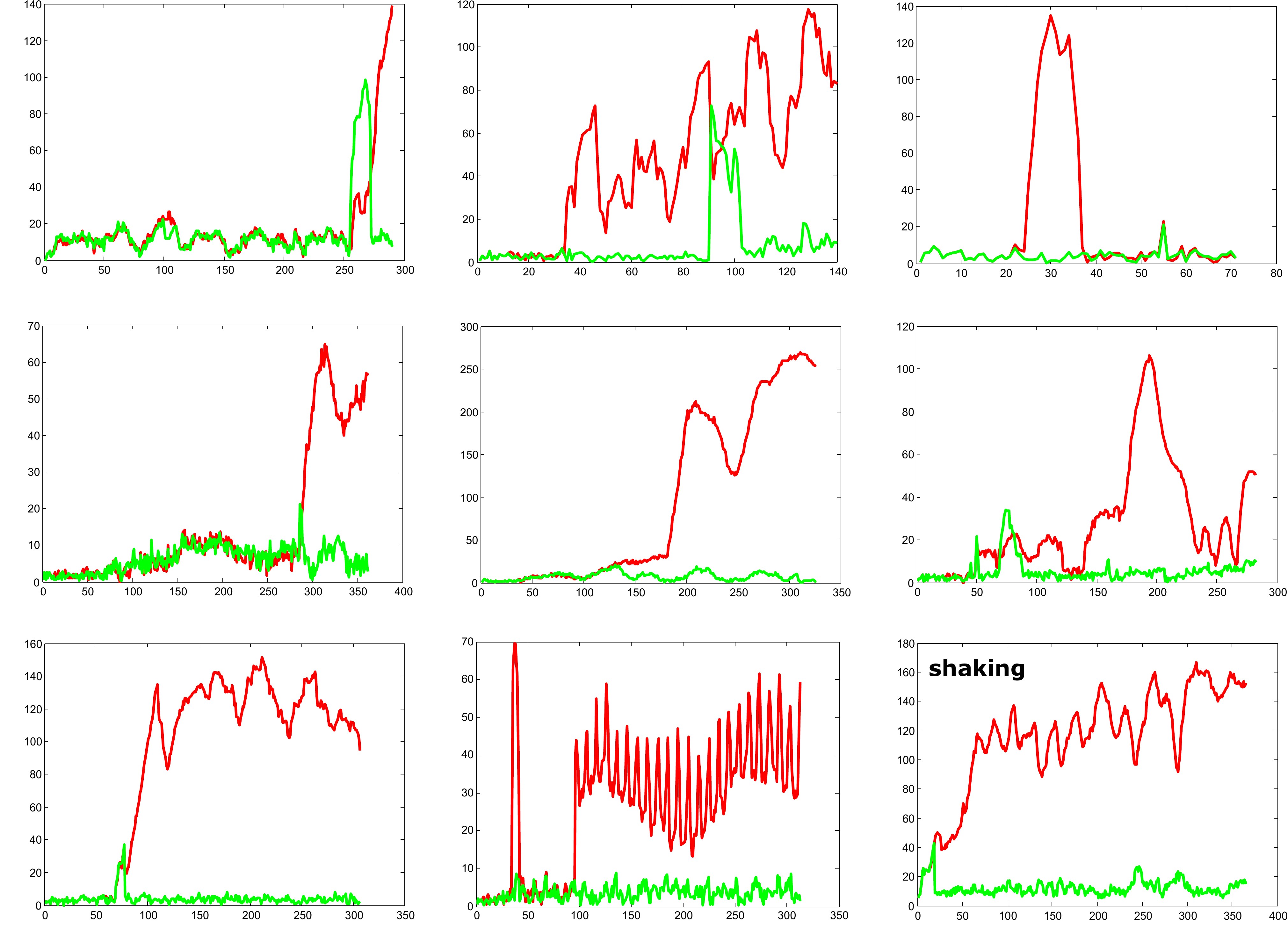}
		\end{center}
		\caption{ The comparison between OCT-KCF(green line) and KCF(red line) based on central location error (CLE). }
		\label{fig:cle}
	\end{figure*}
	
	\begin{figure*} [t!]
		\begin{center}
			\includegraphics[width=0.7\textwidth]{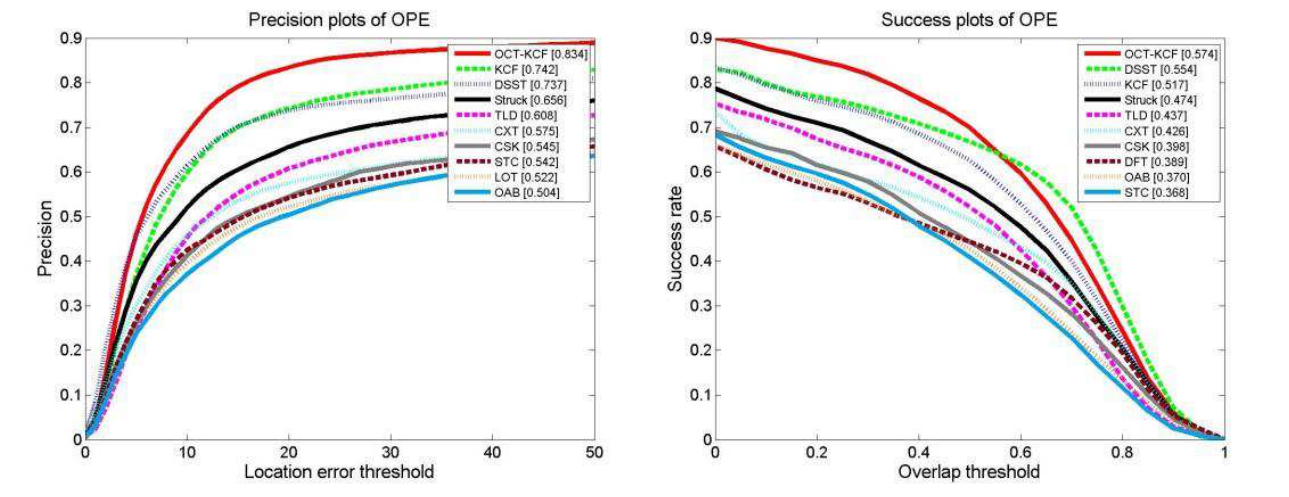}
		\end{center}
		\caption{Success and precision plots according to the online tracking benchmark \cite{dataset} }
		\label{fig:overall}
	\end{figure*}

	Fig. \ref{fig:gaussian} shows that KCF achieves a good performance when the correlation response of the target image follows a Gaussian distribution, \emph{i.e.}, in the basketball sequence. A failure, \emph{i.e.}, in the shaking sequence, is observed when the output is sharply changed. 
	 Fig. \ref{fig:gaussian} also shows that the proposed OCT method can force correlation response of KCF to follow a near-Gaussian distribution, and improves the tracking results of KCF on the shaking sequence. {We compare OCT-KCF with KCF in terms of the central location error (CLE) in Fig. \ref{fig:cle}. It can been seen that the proposed OCT-KCF gets stable performance in terms of CLE, which is clearly indicated by the smooth curves. In contrast, the curves of KCF have hitting turbulence. As illustrated in the coke sequence, both OCT-KCF and KCF lose the target at about the $275^{th}$ frame, nevertheless, the OCT-KCF can relocate it at about the $275^{th}$ frame while KCF fails to do that. The reason is that OCT can help KCF finding the candidate patch whose correlation response satisfies a Gaussian distribution and constraining the tracker from drifting. Similarly, the OCT-KCF tracker achieves much better performance in the sequences of couple, deer, football, etc., than KCF. The CLE results support our previous analysis that OCT-KCF significantly outperforms the conventional KCF.}
	
\begin{figure*} [t!]
			\begin{center}
				\includegraphics[width=1\textwidth]{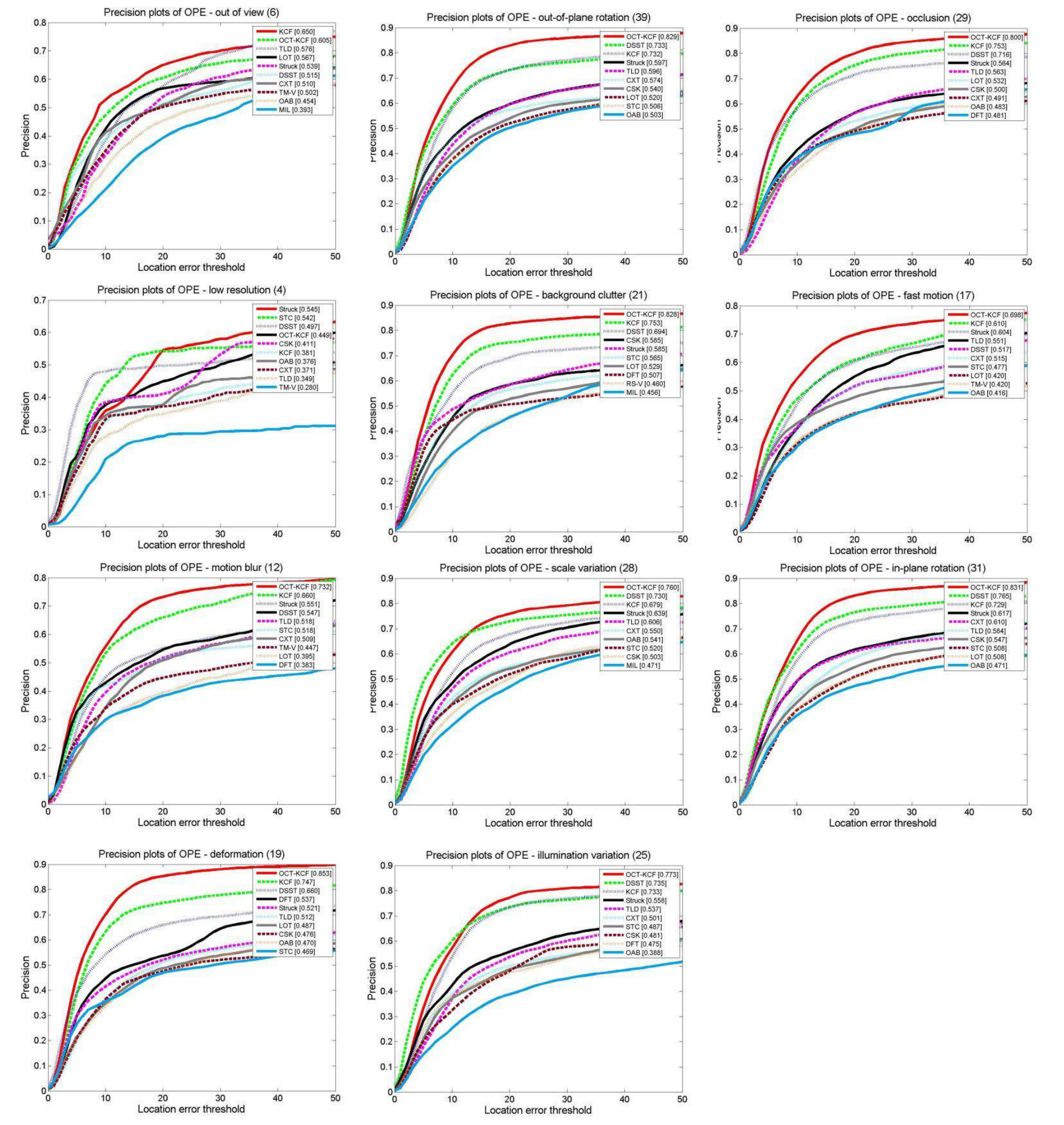}
			\end{center}
			\vspace{-0.1cm}
			\caption{Precision plots for the 11 attributes of the online tracking benchmark.}
			\vspace{-0.2cm}
			\label{fig:pr}
		\end{figure*}

\begin{figure*} [t!]
		\begin{center}
			\includegraphics[width=0.85\textwidth]{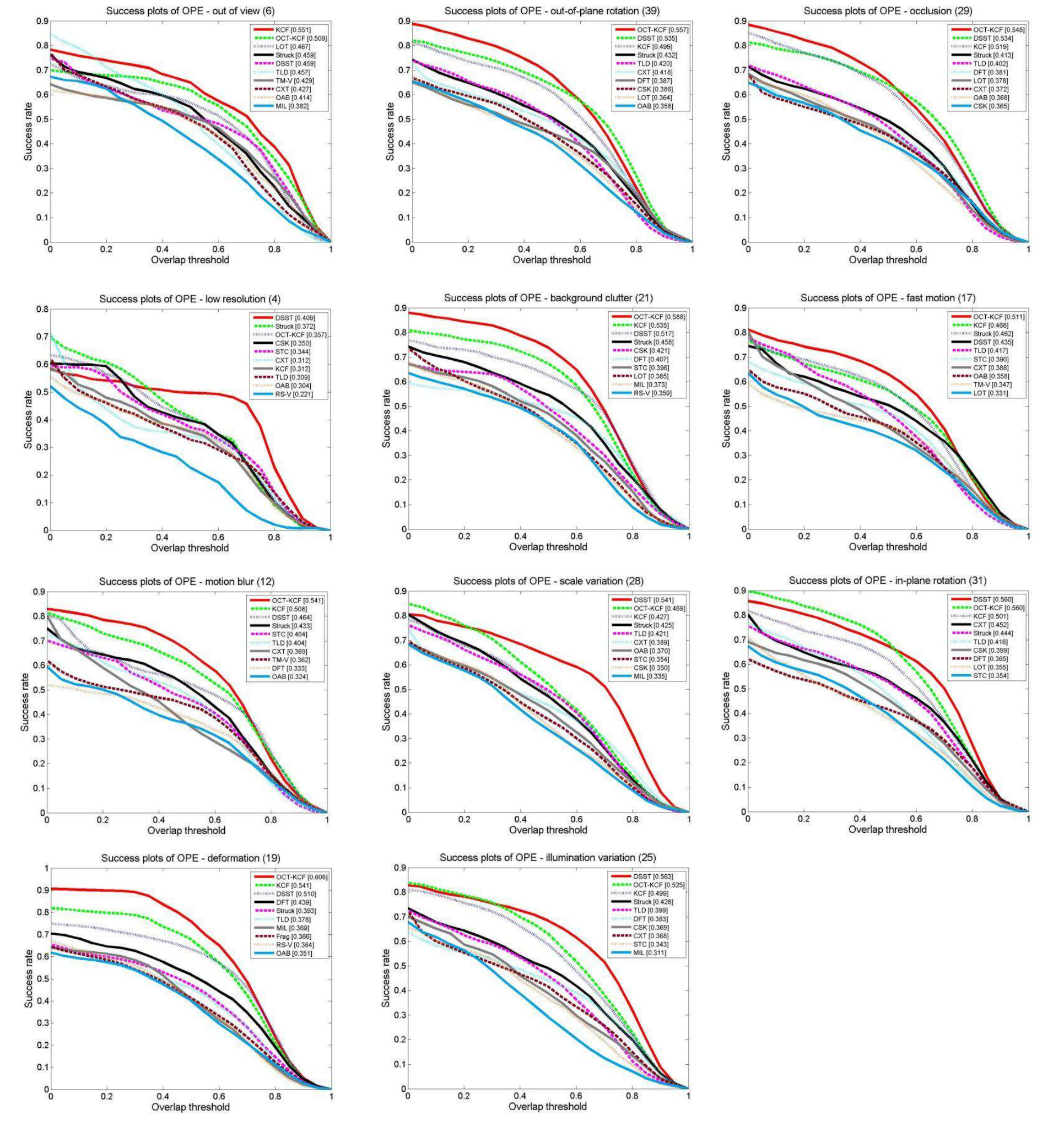}
		\end{center}
		\vspace{-0.1cm}
		\caption{Success plots for the 11 attributes of the online tracking benchmark.}
		\vspace{-0.2cm}
		\label{fig:sr}
	\end{figure*}

	
	{In Fig.\ref{fig:overall}, we report the precision plots which measures the ratio of successful tracking frames whose tracker output is within the given threshold (the x-axis of the plot, in pixels) from the ground-truth, measured by the center distance between bounding boxes. The overall success and precision plots generated by the benchmark toolbox are also reported. These plots report top-10 performing trackers in the benchmarks. As shown in Tab. \ref{tab:tt2}, the proposed method reports the best results. The OCT-KCF and KCF achieve 57.4\% and 51.7\% based on the average success rate, while the famous Struck and TLD trackers respectively achieve 47.4\% and 43.7\%. In terms of Precision, OCT-KCF and KCF respectively achieve 83.4\% and 74.2\% when the threshold is set to 20.  We also compare with DSST, one of latest variants of KCF, which shows that OCT-KCF achieves a significant performance improvement in terms of precision (10.7\%  improved) and success rate (2\%  improved).  These results confirm that the Gaussian prior constraint model contributes to our tracker and enable it performs better than state-of-the-art trackers. The full set of plots generated by the benchmark toolbox are also reported in 	Fig. \ref{fig:sr} and Fig. \ref{fig:pr}
		. From the experimental results, it can be seen that the proposed OCT-KCF achieves significantly higher performance in cases of in-plane rotation (5.9\% improvement over KCF), scale variations (4.2\% improvement over KCF), deformations (6.7\% improvement over KCF), motion blue (3.3\% improvement over KCF) than other trackers (\emph{i.e.}, KCF). This shows that the distribution constrained tracker is more robust to variations mentioned above.}
	
\begin{figure*} [t!]
		\begin{center}
			\includegraphics[width=1\textwidth]{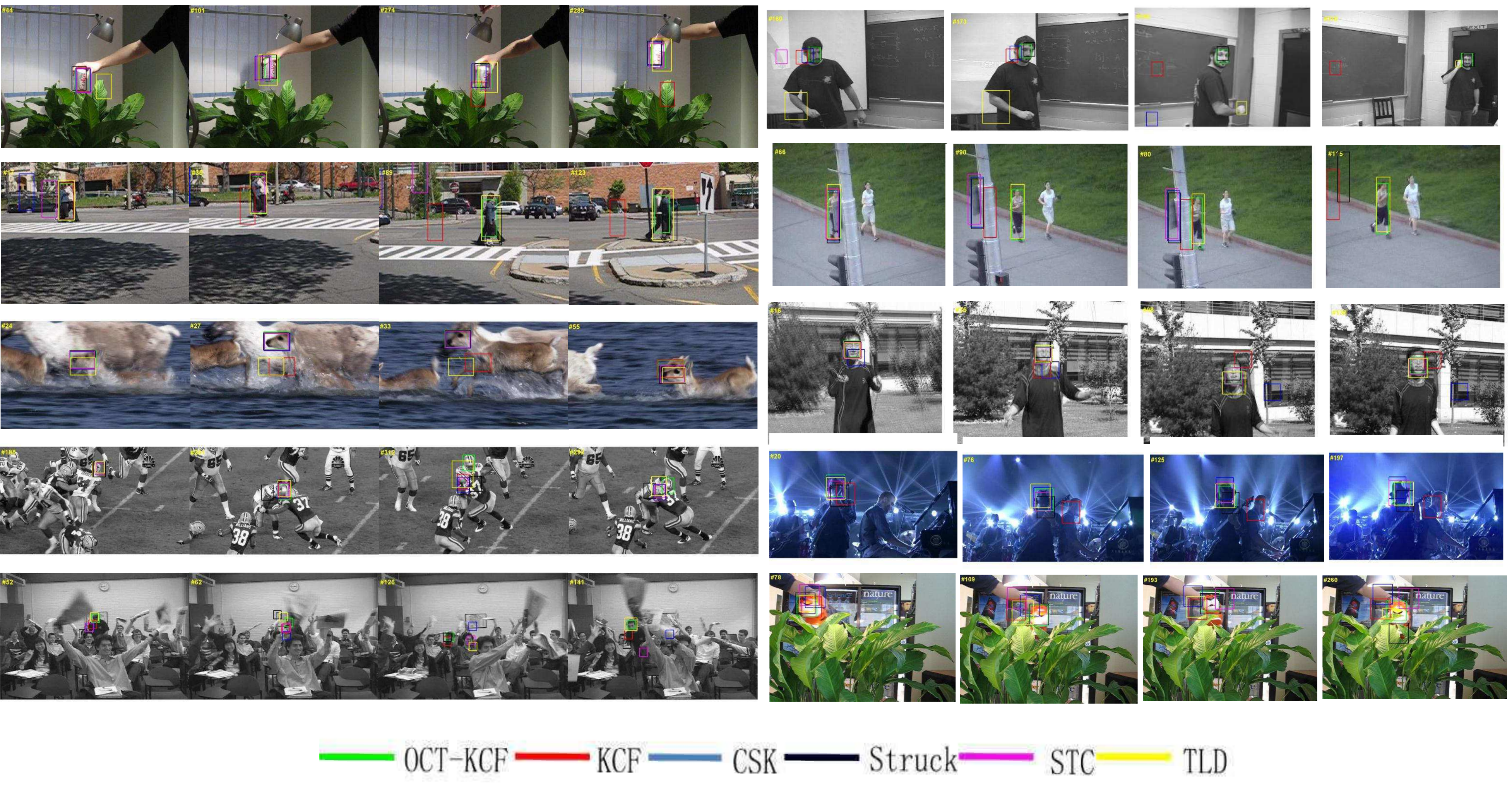}
		\end{center}
		\vspace{-0.1cm}
		\caption{Illustration of some key frames.}
		\vspace{-0.2cm}
		\label{fig:keyframes}
	\end{figure*}

	
	{In Fig. \ref{fig:keyframes}, we illustrate tracking results from some key frames. In the first row, OCT-KCF can precisely track the coke, while the conventional KCF tracker fails to do that. The famous TLD tracker could relocate the coke target after missing it in $44^{th}$ frame. Nevertheless the tracking bounding boxes of the TLD tracker is not as precise as those of OCT-KCF. It is also observed that our proposed OCT-KCF tracker works very well in other sequences, e.g., couple, deer, and football. In contrast, all other compared trackers get false or imprecise results in one sequence at least.}
	{On an Intel I5 3.2 GZ (4 cores) CPU and 8G RAM, the KCF can run up to 185 FPS, while the OCT-KCF achieves 51 FPS. Without losing the real-time performance, the tracking performance is significantly improved by OCT-KCF about 6\% on the average success rate and 10\% on the precision.}
	\begin{table*}[t!]
		\caption{Comparisons with state-of-the-art trackers on the 51 benchmark sequences.}
		\label{tab:tt2}
		\centering
		\begin{tabular}{l|c|c|c|c|c|c|c|}
			& \multicolumn{1}{c}{\rotatebox[origin=c]{0}{\texttt{OCT-KCF}}}&
			\multicolumn{1}{c}{\rotatebox[origin=c]{0}{\texttt{KCF}}} & \multicolumn{1}{c}{\rotatebox[origin=c]{0}{\texttt{DSST}}} & \multicolumn{1}{c}{\rotatebox[origin=c]{0}{\texttt{TLD}}} & \multicolumn{1}{c}{\rotatebox[origin=c]{0}{\texttt{STC}}} & \multicolumn{1}{c}{\rotatebox[origin=c]{0}{\texttt{CSK}}} & \multicolumn{1}{c}{\rotatebox[origin=c]{0}{\texttt{Struck}}}\\
			\hline
			Ref. & Ours &\cite{kcf} & \cite{DSST}&\cite{tld}  &\cite{stc} & \cite{csk} &\cite{struck}  \\
			\hline
			Speed (FPS) & 51 & 185 & 59 &	87 &	410	 & 	430 &13 \\
			Precision& 83.4 & 74.2 & 73.7&	60.8 &	54.2	 &54.5	 &65.6 \\
			Success rate & 57.4 & 51.7 &	55.4&	43.7 &	36.8	 &39.8	 & 47.4 \\
			\hline

		\end{tabular}
	\end{table*}

	\subsection{Conclusion}
	We proposed an output constraint transfer (OCT) method
	to enhance commonly used correlation filter for object
	tracking. OCT is a new framework introduced to improve
	the tracking performance based on the Bayesian optimization method. To improve the robustness of the correlation
	filter to the variations of the target, the correlation response (output) of the test image is reasonably considered
	to follow a Gaussian distribution, which is theoretically
	transferred to be a constraint condition in the Bayesian
	optimization problem, and successfully used to solve the
	drifting problem. We obtained a new theory which can
	transfer the data distribution to be a constraint of an optimization problem, which leads to an efficient framework
	to calculate correlation filter. Extensive experiments and
	comparisons on the tracking benchmark show that the
	proposed method significantly improved the performance
	of KCF, and achieved a better performance than state-of-
	the-art trackers. In addition, the performance is obtained
	without losing the real-time tracking performance. \textbf{Although high performance is obtained, the drifting detection function (Equ. \ref{drifting}) is too simple for practical tracking problems, which might fail to start the fine-tunning process when the targets suffer from occlusion or abrupt motion. Therefore, the future work will focus on new drifting detection methods to achieve higher tracking performance. Moreover, we will also try to improve OCT based on other machine learning methods, such as \cite{nanda, nanda1,shenda,shenda1,shenda2}, to solve the long-term tracking problem.}
	
\subsection{Acknowledgment}
This work was supported in part by the Natural Science Foundation of China  under Contract 61672079,  61672357 and by the Science and Technology Innovation Commission of Shenzhen under Grant JCYJ20160422144110140.  The work of B. Zhang was supported by the Program for New Century Excellent Talents University within the Ministry of Education,  China, and Beijing Municipal Science and Technology Commission Z161100001616005.

\end{document}